%% file: main.tex
\documentclass[10pt,twocolumn,letterpaper]{article}

\usepackage[pagenumbers]{cvpr} 
\usepackage{algorithm}
\usepackage{algpseudocode}
\usepackage{tipa}
\usepackage{wrapfig}
\usepackage{adjustbox}
\usepackage{multicol}
\usepackage{multirow}
\usepackage{setspace}
\usepackage{amsbsy}
\usepackage{placeins}
\usepackage{colortbl}
\usepackage{mathtools}      

\usepackage{amsmath,amssymb,bm}
\usepackage{graphicx}
\usepackage{booktabs}
\usepackage{array}

\input{preamble}
\definecolor{cvprblue}{rgb}{0.21,0.49,0.74}
\usepackage[pagebackref,breaklinks,colorlinks,allcolors=cvprblue]{hyperref}

\def\paperID{} 
\def\confName{CVPR}
\def\confYear{2026}

\title{AvatarForcing: One-Step Streaming Talking Avatars via Local-Future Sliding-Window Denoising}

\author{
Liyuan Cui$^{1,3}$\footnotemark[1] \footnotemark[2] \quad
Wentao Hu$^{2,3}$\footnotemark[1] \quad
Wenyuan Zhang$^{4}$\footnotemark[1] \quad
Zesong Yang$^{1}$ \quad
Fan Shi$^{3}$ \quad
Xiaoqiang Liu$^{3}$ \\
\vspace{5pt} \\
$^{1}$Zhejiang University \quad
$^{2}$ Beijing University of Posts and Telecommunications \quad
\vspace{5pt} \\
$^{3}$Kling Team, Kuaishou Technology \quad
$^{4}$Tsinghua University \quad
\vspace{5pt} \\
{\tt\small Project Page: \url{https://cuiliyuan121.github.io/AvatarForcing/}}
}

\begin{document}

\twocolumn[{%
\renewcommand\twocolumn[1][]{#1}%
\maketitle
\vspace{-5mm}
\includegraphics[width=\linewidth]{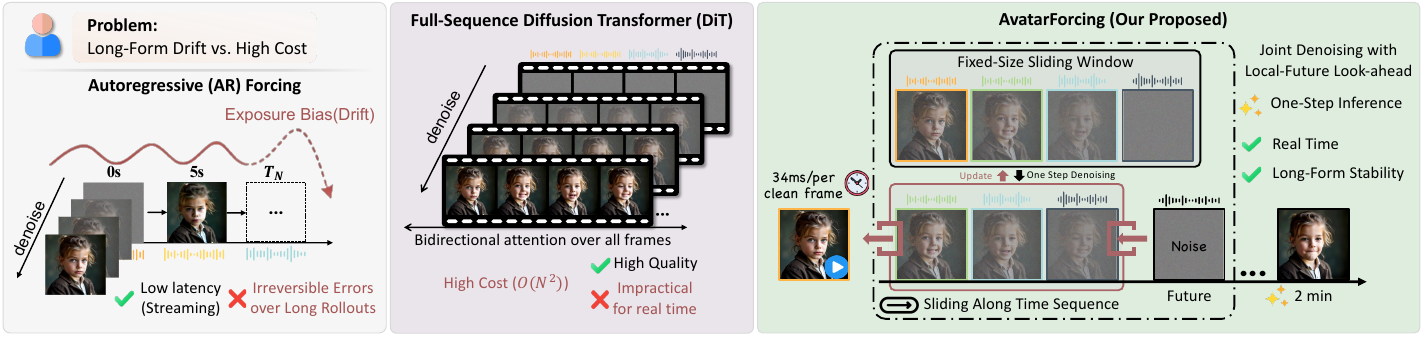}
\vspace{-5mm}
\captionof{figure}{
    \textbf{AR forcing vs.\ full-sequence DiT vs.\ AvatarForcing.} AvatarForcing enables real-time, long-form talking-avatar generation from a reference image and streaming audio. It performs one-step joint denoising in a fixed sliding window to introduce bounded local-future context at constant latency, reducing autoregressive drift without full-sequence diffusion (34\,ms/frame).
}
\vspace{1em}
\label{fig:teaser}
}]

\renewcommand{\thefootnote}{\fnsymbol{footnote}}
\footnotetext[1]{Equal contribution.}
\footnotetext[2]{This work was conducted during the author’s internship at Kling Team, Kuaishou Technology.}

\input{sec/0_abstract}    
\input{sec/1_intro}
\input{sec/2_related}
\input{sec/3_method}
\input{sec/4_experiment}

\input{sec/5_conclusion}
{
    \small
    \bibliographystyle{IEEEtran}
    \bibliography{main}
}


\end{document}

%% file: sec/0_abstract.tex
\begin{abstract}
Real-time talking avatar generation requires low latency and minute-level temporal stability. Autoregressive (AR) forcing enables streaming inference but suffers from exposure bias, which causes errors to accumulate and become irreversible over long rollouts. In contrast, full-sequence diffusion transformers mitigate drift but remain computationally prohibitive for real-time long-form synthesis. We present AvatarForcing, a one-step streaming diffusion framework that denoises a fixed local-future window with heterogeneous noise levels and emits one clean block per step under constant per-step cost. To stabilize unbounded streams, the method introduces dual-anchor temporal forcing: a style anchor that re-indexes RoPE to maintain a fixed relative position with respect to the active window and applies anchor-audio zero-padding, and a temporal anchor that reuses recently emitted clean blocks to ensure smooth transitions. Real-time one-step inference is enabled by two-stage streaming distillation with offline ODE backfill and distribution matching. Experiments on standard benchmarks and a new 400-video long-form benchmark show strong visual quality and lip synchronization at 34\,ms/frame using a 1.3B-parameter student model for real-time streaming.
\end{abstract}

%% file: sec/1_intro.tex
\section{Introduction}
\label{sec:intro}

Talking-avatar video synthesis is a central problem in digital human research, with applications in virtual communication, content creation, and embodied AI. Diffusion Transformer (DiT) models conditioned on audio~\cite{tian2025emo2,ding2025kling,gao2025wan}, text prompts~\cite{wan2025basemodel,zhang2025waver,gao2025seedance}, and portrait inputs~\cite{liu2024anitalker,zhang2025xactro} have improved fidelity and controllability for short clips. Real-time long-form streaming, however, requires constant per-frame latency and stable appearance over minutes, without identity or color drift.

These requirements expose a fundamental tension in DiT attention. Full-sequence bidirectional attention denoises all frames at every step, which makes self-attention computationally expensive and incompatible with constant-latency streaming. Many diffusion pipelines also assume that control signals are available upfront, which limits interactive updates~\cite{gu2025diffusion,wang2024motionctrl}. Autoregressive (AR) denoising supports streaming by predicting frames sequentially under causal conditioning~\cite{chen2025midas,casvid,teng2025magi,deng2024autoregressive}. Strict causality, however, introduces exposure bias: once a frame is emitted, it becomes fixed context, and small appearance or motion errors can accumulate over thousands of steps, leading to drift and flicker~\cite{low2025talkingmachines,yang2025longlive}. Inference-aware training such as self-forcing~\cite{huang2025selfforcing} reduces the train--test gap, and teacher-guided variants use a bidirectional teacher to mitigate drift~\cite{cui2025selfforcingpp,chen2026contextforcing}. Yet inference remains strictly causal, with limited look-ahead and no reliable mechanism to re-anchor identity during unbounded generation.

AvatarForcing addresses this limitation by introducing bounded look-ahead while maintaining constant latency. Instead of denoising a single block under strictly causal context, the method maintains a fixed window of latent blocks at heterogeneous noise levels and applies one joint denoising update to the entire window at each step. The leftmost block is emitted, the window shifts, and fresh noise is appended. This design is important for long rollouts in two ways. First, the current frontier can condition on a small amount of future context; although future blocks are noisy, they still provide coarse cues about motion and structure. Second, each block is revisited multiple times while it traverses the window, before it is committed to history.

It is useful to view streaming diffusion as allocating a fixed latency budget between denoising depth (denoising passes per step, $N$) and bounded look-ahead (window length, $L$). Forcing-style causal methods use $L=1$ and rely on larger $N$ for refinement. AvatarForcing sets $N=1$ and increases $L$ to introduce bounded look-ahead, while still refining each block multiple times as it moves through the window. We formalize this mechanism as $\mathcal{B}_{L,N}$ (Sec.~\ref{sec:method}) and study the roles of $L$ and $N$ in Sec.~\ref{sec:exp}. Under matched compute $L\cdot N$, increasing $L$ often improves long-horizon stability more than increasing $N$, which suggests that limited look-ahead is more effective than repeatedly refining a short context.

Bounded look-ahead alone does not prevent minute-scale identity and color drift, because emitted frames cannot be revised. We therefore introduce a dual-anchor KV cache that provides both a stable identity reference and short-term temporal continuity. The style anchor maintains a reference frame as a persistent appearance cue: keys are stored in pre-RoPE space and re-applied with RoPE at each step using a fixed offset $d$ relative to the current window (default $d=-1$), so the anchor remains aligned as the stream grows. The temporal anchor caches recently emitted clean blocks to preserve short-term dynamics and smooth transitions across window boundaries. For audio-driven synthesis, per-frame audio tokens are aligned to the active window, and anchor frames use zero audio features so that identity anchoring does not inject stale speech content.

To enable real-time one-step inference, we distill a global bidirectional teacher using a two-stage training procedure. Offline ODE backfill records teacher trajectories and supports audio-conditioned ODE regression pre-training that maps intermediate noise levels to the clean endpoint in one step. We then apply DMD-based distribution matching on rolling-window student simulations, which adapts the denoiser to heterogeneous-noise windows without multi-step teacher sampling in the loop. The resulting 1.3B-parameter student runs at 34\,ms/frame while generating minute-long videos with stable identity and accurate lip synchronization.

We evaluate AvatarForcing on diverse benchmarks and introduce a new 400-video long-form benchmark. We report both short-clip quality metrics and long-horizon stability measures.

Our main contributions are summarized as
follows:
\begin{itemize}
\item We introduce $\mathcal{B}_{L,N}$, a one-step streaming diffusion mechanism that denoises a fixed local-future window with heterogeneous noise levels. The mechanism provides bounded look-ahead under constant per-step compute, and it implicitly refines each block multiple times before emission.
\item We propose a dual-anchor KV cache for unbounded streaming. A RoPE re-indexed style anchor stabilizes identity under growing absolute time, and a temporal anchor reuses recent clean blocks to maintain smooth transitions; anchor-audio zeroing prevents the anchor path from carrying stale speech content.
\item We develop a two-stage distillation pipeline for one-step inference. Offline ODE backfill enables audio-conditioned ODE regression pre-training, and DMD post-training adapts the student to its rolling-window rollout distribution without multi-step teacher sampling in the loop.
\item We construct a 400-video long-form benchmark and show strong quality, stability, and lip synchronization under real-time streaming with a 1.3B-parameter student model running at 34\,ms/frame.
\end{itemize}

%% file: sec/2_related.tex
\section{Related Work}
\label{sec:related_works}
\subsection{Diffusion Models for Avatar Generation}
Recent diffusion-based methods achieve strong visual quality for short-form digital human synthesis~\cite{wang2025fantasytalking, ding2025kling, hello3, chen2025hunyuanvideo, tian2024emo, tian2025emo2,lin2025omnihuman1,jiang2025omnihuman1.5}. However, the use of fixed-window bidirectional attention makes constant-latency streaming difficult and limits minute-scale stability. StreamAvatar~\cite{sun2025streamavatar} studies real-time interactive conversational avatars. AvatarForcing targets one-step streaming talking-avatar synthesis by denoising a bounded local-future window and introducing dual-anchor stabilization. Frame packing~\cite{gao2025wan} and overlapping-window inference~\cite{chen2025hunyuanvideo} extend the effective horizon and improve efficiency, but unbounded streaming remains challenging without explicit stabilization.

\subsection{Autoregressive Video Generation}
Autoregressive video generation extends fixed-window diffusion models to long-horizon synthesis by producing frames sequentially under causal conditioning~\cite{casvid, zhang2025xactro, sun2025ardiffusion, cheng2025playing, gao2024ca2, xu2024msc}. Hybrid schemes combine autoregressive rollouts with diffusion-based updates to improve short-term stability~\cite{chen2024diffusionforcing, guo2025long}. Inference-aware training further reduces train--test mismatch by exposing a model to its own rollouts during training~\cite{huang2025selfforcing,yang2025longlive}. Several methods~\cite{cui2025selfforcingpp,chen2026contextforcing} mitigate drift by applying teacher feedback to student rollouts. Causal Forcing~\cite{zhu2026causalforcing} analyzes the gap in autoregressive distillation and refines ODE-based initialization using an autoregressive teacher. These methods remain strictly causal, which limits look-ahead and allows residual errors to accumulate over long rollouts.
\looseness=-1
AvatarForcing relaxes strict causality by enabling bounded look-ahead within a fixed window and uses a dual-anchor KV cache to stabilize long-term identity and short-term dynamics under unbounded streaming.

\subsection{Efficiency and Consistency Enhancements}
Several works improve long-form efficiency through streaming-aware and memory-efficient diffusion inference~\cite{chen2024streaming,zheng2024memo,fan2025syncdiff,li2025ditto}. FIFO-Diffusion~\cite{kim2024fifodiffusion} applies diagonal denoising over a FIFO queue to achieve constant memory usage, whereas StreamDiT~\cite{kodaira2025streamdit} uses sliding windows with recurrent KV caching to avoid full-sequence attention. LiveAvatar~\cite{huang2025liveavatar} focuses on system-level acceleration by using timestep-forcing pipeline parallelism to distribute multi-step denoising across GPUs for low-latency streaming. LongLive~\cite{yang2025longlive} reuses buffered latents to improve short-term continuity during rollout. Although these methods reduce computational cost and enable streaming, persistent appearance anchoring and frame-synchronous conditioning under bounded context remain open challenges.

%% file: sec/3_method.tex
\begin{figure*}[t]
\centering
\includegraphics[width=\linewidth]{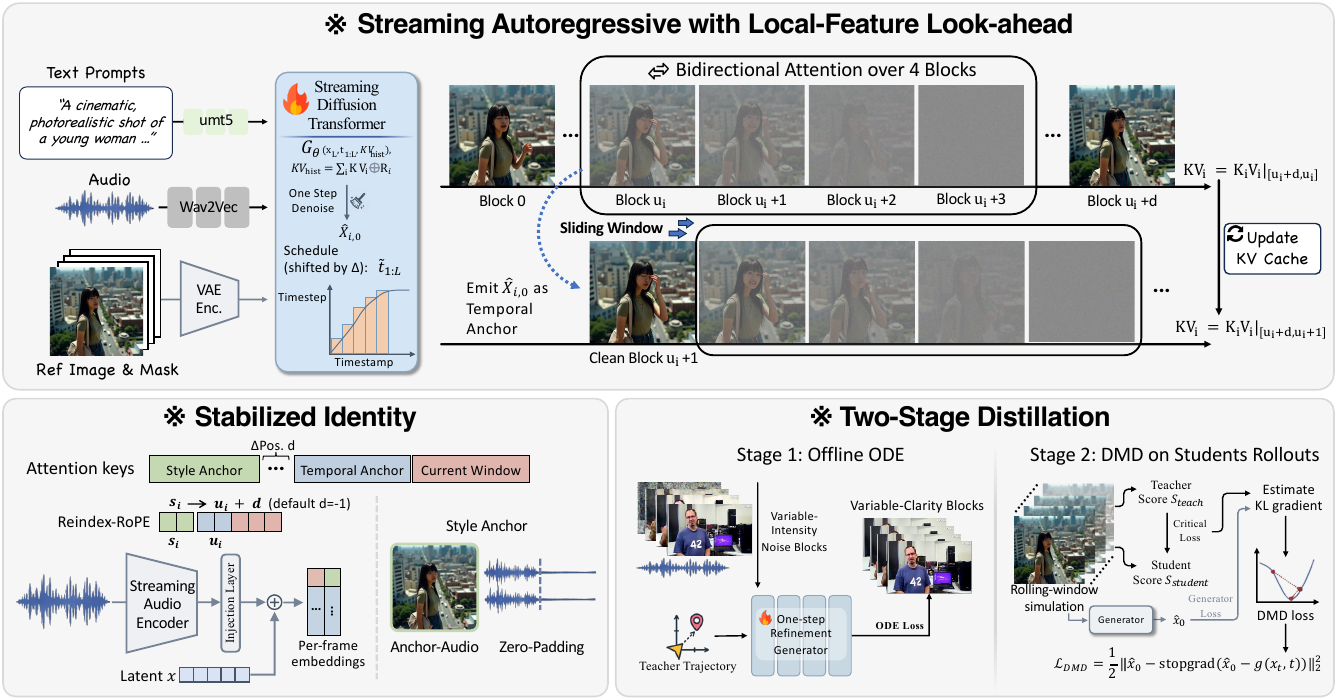}
\caption{\textbf{Windowed denoising with dual anchors and two-stage distillation.} AvatarForcing performs one-step denoising over a fixed local-future window with heterogeneous timesteps (cleaner on the left and noisier on the right). At each step, the window is jointly updated under bidirectional attention to emit the leftmost clean block, slide the window, and append fresh noise. Long-horizon stability is supported by a dual-anchor KV cache, including a RoPE re-indexed style anchor (with anchor-audio zero-padding) and a temporal anchor constructed from recent clean blocks. Real-time one-step inference is achieved by distilling a global bidirectional teacher into a streaming student via two-stage training with offline ODE backfill and DMD post-training on student rollouts.}
\label{fig:pipeline_overall}
\end{figure*}

\section{Method}
\label{sec:method}

This work studies long-horizon, audio-driven talking-avatar synthesis from a reference image and control signals such as audio and text. The objective is to achieve real-time latency while maintaining stable identity and temporally consistent motion. AvatarForcing is a one-step streaming diffusion mechanism: at each step, it jointly denoises a fixed local-future window and emits one clean block. Long-horizon stability is supported by dual-anchor KV caching with RoPE re-indexing and anchor-audio zero-padding, and two-stage streaming distillation transfers supervision from a bidirectional teacher.
Fig.~\ref{fig:pipeline_overall} provides an overview.

\paragraph{Notation and mechanism \texorpdfstring{$\mathcal{B}_{L,N}$}{B_L,N}.}
We group $B$ consecutive latent frames into a block (the KV-cache granularity) and maintain an $L$-block sliding window ($B=4$ unless stated otherwise).
The mechanism is denoted by $\mathcal{B}_{L,N}$: at each step, $N$ joint denoising passes are applied to an $L$-block window to emit one clean block, yielding $L\cdot N$ denoising updates per emitted block.
Throughout the paper, one-step refers to the streaming setting with $N=1$, \ie, a single joint denoising update per streaming step.
In the default real-time setting, we use $N=1$ and select $L$ to balance stability and latency (Sec.~\ref{sec:exp}); each block remains in the window for $L$ steps and is refined multiple times before emission.

\subsection{Rolling-Window Sequential Denoising}

Existing autoregressive diffusion methods often adopt single-frame causal denoising~\cite{sun2025ardiffusion,casvid,gu2025far}, where the current frame is denoised and then committed to history as immutable context. This design is efficient, but it implicitly assumes that locally optimal per-frame predictions remain globally consistent over long rollouts. In practice, small errors in appearance, geometry, or motion accumulate over time, and the model cannot revise frames after emission.

To preserve streaming emission while enabling bounded joint refinement near the generation frontier, AvatarForcing introduces a sliding-window joint denoising mechanism with bounded look-ahead. Concretely, at streaming step $i$, the method maintains an $L$-block window
\begin{equation}
X^{i} = \{ \mathbf{x}_{i}^{t_1}, \mathbf{x}_{i+1}^{t_2}, \dots, \mathbf{x}_{i+L-1}^{t_L} \},
\end{equation}
where each block $\mathbf{x}_{k}\in\mathbb{R}^{B\times C\times H\times W}$ contains $B$ consecutive latent frames, and $t_1<\cdots<t_L$ are heterogeneous noise levels (cleaner on the left and noisier on the right).
We follow the standard diffusion convention where timestep $0$ denotes the clean endpoint and larger timesteps correspond to higher noise levels. Therefore, $t_1$ is closest to the clean endpoint and $t_L$ is the noisiest stage in the window. At each streaming step, the scheduler reduces the noise level of each block by one stage (or by $N$ sub-steps) while preserving the within-window ordering: blocks shift left toward smaller timesteps, and a fresh noisy block is appended on the right at timestep $t_L$. This creates a graded refinement band: left blocks are near clean and require fine corrections, whereas right blocks remain noisy and allow larger structural updates.

\paragraph{Local-future guidance under bounded look-ahead.}
Within each window, windowed bidirectional self-attention is applied over the entire $L$-block buffer so that denoising at the current frontier can condition on a bounded set of future (yet still noisy) blocks.
Concretely, the prediction of the emitted block $\widehat{\mathbf{x}}_{i,0}$ can attend to all tokens in $\{\mathbf{x}_{i}^{t_1},\dots,\mathbf{x}_{i+L-1}^{t_L}\}$, including future blocks that remain noisy latents. As a result, look-ahead is explicitly bounded to $(L-1)$ blocks. This provides bounded look-ahead that is unavailable in strictly causal AR forcing while remaining compatible with streaming, because audio conditioning is aligned to the same window.
In streaming talking-avatar synthesis, this implies an algorithmic audio look-ahead of $(L-1)B$ frames; we report its time scale under the default configuration in Sec.~\ref{sec:exp}.

\paragraph{Why noisy-future correction is more effective than repeated sampling.}
Under a fixed latency budget, increasing the number of denoising passes $N$ refines the same window multiple times but does not expand the set of frames available for conditioning. From a probabilistic perspective, a one-step denoiser can be viewed as approximating a conditional estimator of the clean signal given the noisy inputs and the available context. Repeating this refinement does not resolve the intrinsic ambiguity induced by missing context, and it can compound systematic bias from the causal history, which often manifests as over-smoothed motion. In contrast, increasing the window length $L$ introduces forward references: the current frontier is denoised while conditioning on future blocks at higher noise levels. Although these blocks are noisy, they carry coarse structure and motion cues that constrain the update direction and reduce boundary discontinuities before emission. This mechanism provides a stronger form of error correction than additional passes on a short window and helps explain why enlarging $L$ is often more beneficial than enlarging $N$ under matched compute $L\cdot N$ (Sec.~\ref{sec:exp_lvsn}).

During a forward pass, the student predicts the clean targets for the entire window:
\begin{equation}
\widehat{X}^{i}_0
= G_{\theta}\left(
X^{i},
t_{1:L},
\mathrm{KV}_{\text{hist}},
a_{i:i+L-1}\right),
\end{equation}
where $a_{i:i+L-1}$ denotes per-frame audio aligned to the window.

After $N$ joint denoising passes (that is, $\mathcal{B}_{L,N}$), the method emits the leftmost block $\widehat{\mathbf{x}}_{i,0}$ as the next output. A new latent block sampled from the prior
$x_{i+L}^{t_L} \sim \mathcal{N}(0,I)$
is appended to the noisy end.
Thus, the next window becomes
\begin{equation}
X^{i+1}
= \{
\widehat{\mathbf{x}}_{i+1}^{t_1},
\widehat{\mathbf{x}}_{i+2}^{t_2},
\dots,
\widehat{\mathbf{x}}_{i+L-1}^{t_{L-1}},
\mathbf{x}_{i+L}^{t_L}
\}.
\end{equation}

This formulation enables mutual refinement within a local window because future noisy blocks impose additional constraints on the current frontier. It also provides implicit multi-pass correction: as a block traverses the window, the block is revisited for approximately $L \cdot N$ updates before emission, which mitigates long-horizon error accumulation. Finally, the per-step cost depends on the fixed window and caches rather than on the total video length. An algorithmic summary of online inference is provided in Alg.~\ref{alg:inference}.

\begin{algorithm}[t]
\caption{Streaming inference with AvatarForcing ($\mathcal{B}_{L,N}$).}
\label{alg:inference}
{\footnotesize
\algrenewcommand\algorithmicindent{0.9em}
\begin{algorithmic}
\setstretch{1.05}
\State Initialize $X^{1}$ and $\mathrm{KV}\leftarrow\varnothing$.
\For{$i=1,2,\dots$}
\State Compute window audio features $a$; set anchor audio to zero.
\State Re-index style-anchor RoPE and assemble attention keys.
\For{$n=1$ to $N$}
\State $\widehat{X}^{i}_0 \gets G_{\theta}(X^{i}, t_{1:L}, \mathrm{KV}, a)$.
\State Update $(X^{i}, t_{1:L})$ using the scheduler (toward smaller timesteps).
\EndFor
\State Emit $\widehat{\mathbf{x}}_{i,0}$ and update the temporal KV cache.
\State Slide the window and append a fresh noise block.
\EndFor
\end{algorithmic}}
\end{algorithm}

\subsection{Dual-Anchor KV Caching with Style Anchor}
\label{sec:kv}
Although the rolling window provides strong short-range refinement, the receptive field remains bounded. Once frames leave the window, the method cannot revisit them, which makes the system vulnerable to slow drift in identity, color tone, and expression style. Relying only on local context forces long-term consistency to be maintained from a short temporal horizon.
To address this issue, we augment sliding-window denoising with a bounded KV cache that maintains two anchors: a global style anchor and a rolling temporal anchor. At each step, attention keys are assembled as [style anchor \textbar{} temporal cache \textbar{} current window], and the temporal cache length is capped by a fixed token budget to keep per-step compute constant.

\paragraph{Temporal anchor (recent clean KV).}
After emitting the next clean block $\widehat{\mathbf{x}}_{i,0}$, the method writes the corresponding KV states into a rolling cache and retains only the most recent clean tokens under the budget. This temporal anchor stabilizes short-term dynamics, for example, head motion and lip trajectories, and suppresses boundary flicker when the window slides.

\paragraph{Style anchor with RoPE re-indexing.}
We keep a reference frame as a persistent style anchor to preserve identity and appearance. The keys are cached in pre-RoPE space and RoPE is applied on the fly so that the anchor stays at a fixed relative position with respect to the current attention context. Let $u_i$ denote the absolute index of the first non-anchor frame in the current attention context (that is, the first frame of [temporal cache \textbar{} current window] at step $i$). We assign the style anchor a virtual index $u_i + d$, where $d$ is a fixed offset (default $d=-1$, hence $u_i + d = u_i - 1$), and compute
\[
\widetilde{\mathbf K}_{\text{anc}}(i) = \mathrm{RoPE}\left(\mathbf K_{\text{anc}}^{\text{pre}},\, u_i + d\right).
\]
This RoPE re-indexing schedule prevents phase mismatch as $u_i$ grows and turns the reference frame into a persistent style anchor rather than a temporally distant frame. Combined with anchor-audio zeroing (Sec.~\ref{sec:audio}), the anchor provides identity guidance without injecting stale speech content.

\subsection{Non-intrusive Per-Frame Audio Injection}
\label{sec:audio}

In streaming with sliding windows and non-consecutive visual frames, conditioning on continuous audio features can introduce cross-window interference that leads to lip jitter and phoneme misalignment. To preserve compatibility with existing T2V backbones, we use lightweight per-frame audio injection that respects window alignment.

\paragraph{Streaming audio encoder.}
Raw audio is processed by a pretrained streaming speech encoder to produce per-frame embeddings:
\begin{equation}
a_t = \mathrm{Enc}_{\text{audio}}(\text{audio segment at time } t).
\end{equation}

\paragraph{Zero-padding anchor.}
Early anchor frames are reused throughout streaming to stabilize identity and appearance. Conditioning anchors on their original audio can misalign speech content with the current window, which introduces spurious correlations and destabilizes lip motion. To avoid this interference, anchor audio features are set to zero, $a^{\text{anchor}} = \mathbf{0}$. The anchors then serve as visual identity references, while the current window provides active audio control.

\paragraph{Non-intrusive fusion.}
Audio is injected through additive modulation of latent tokens in selected transformer layers. For each frame, the audio embedding is projected to the model dimension, broadcast to the latent spatial grid, and added to the patch tokens:
\[
x \leftarrow x + f_{\text{inj}}(a_t).
\]
This design preserves the attention structure (no additional cross-attention blocks) while providing frame-synchronous audio control.

\begin{figure*}[!t]
\centering
\includegraphics[width=\linewidth]{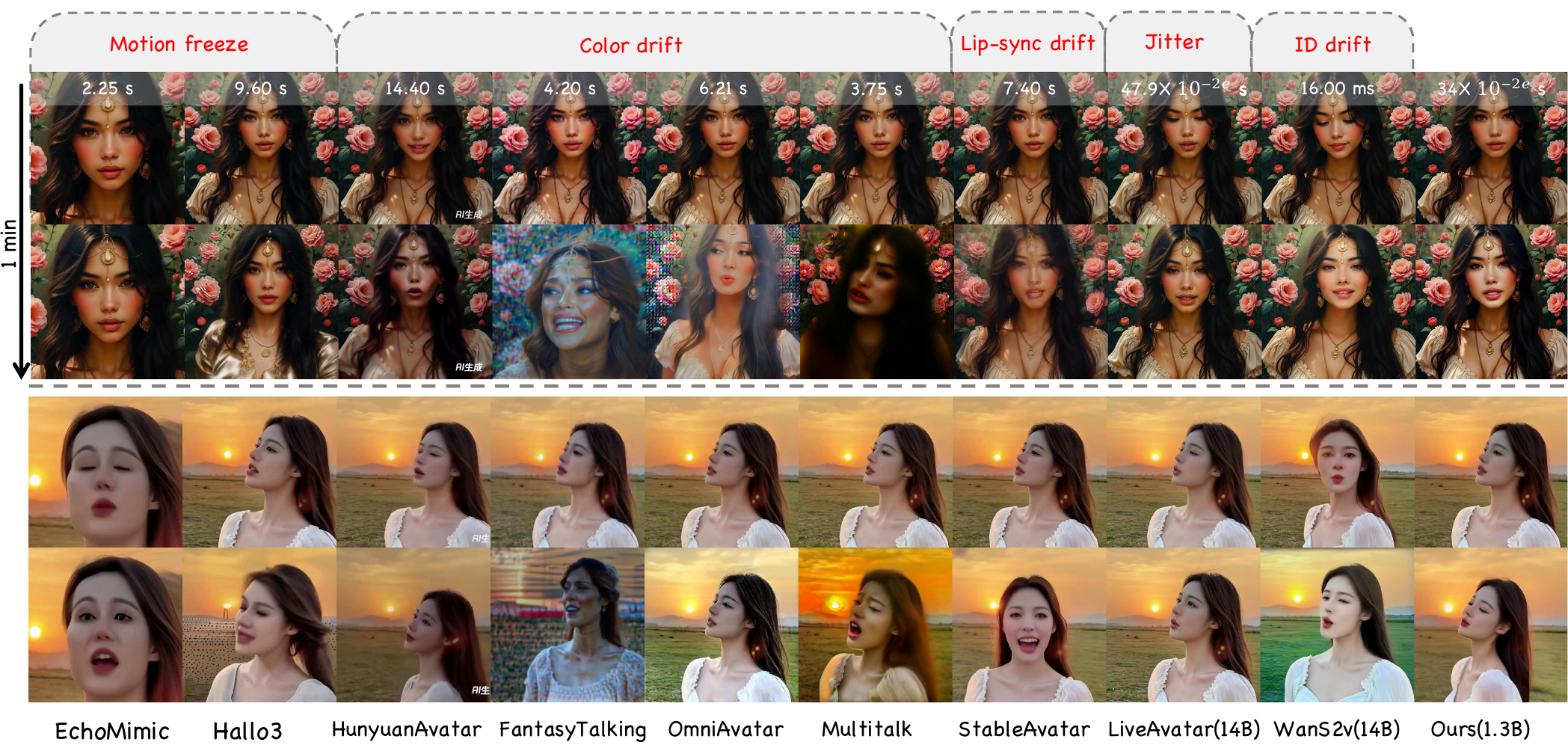}
\caption{\textbf{Long-form qualitative comparison.} We compare AvatarForcing with representative autoregressive and diffusion-based talking-avatar models on long-form generation, with an emphasis on temporal stability, identity preservation, and audio--visual synchronization.}
\label{fig:exp}
\end{figure*}

\definecolor{bestgreen}{HTML}{81C784}
\definecolor{secondgreen}{HTML}{C8E6C9}
\newcommand{\bestcell}[1]{\cellcolor{bestgreen}#1}
\newcommand{\secondcell}[1]{\cellcolor{secondgreen}#1}

\begin{table*}[t]
\centering
\caption{\textbf{Quantitative comparison on CelebV-HQ and long-form videos.} Values are reported as CelebV-HQ\,/\,long-form, except latency (s/frame). CelebV-HQ clips average 5 seconds and long-form videos average 2 minutes. Best and second-best long-form results are highlighted in deep and light green. For pipelined multi-GPU systems (e.g., LiveAvatar~\cite{huang2025liveavatar}), latency (s/frame) denotes the steady-state output interval (approximately $1/\mathrm{FPS}$).}
\label{tab:main_quantitative}
\resizebox{\linewidth}{!}{
\begin{tabular}{lcccccc}
\toprule
Model & FID$\downarrow$ & FVD$\downarrow$ & CSIM$\uparrow$ & Sync-C$\uparrow$ & Sync-D$\downarrow$ & Latency (s/frame)$\downarrow$ \\
\midrule
EchoMimic~\cite{chen2025echomimic}           & 63.26 / 147.13 & 1115.86 / 2102.42 & 0.69 / 0.79 & 4.72 / 3.44 & 9.55 / 10.91 & 2.25 \\
Hallo3~\cite{cui2025hallo3}                  & 47.40 / 122.64 & 488.50 / 1371.98  & 0.85 / 0.83 & 2.67 / 3.90  & 10.29 / 10.42 & 9.60 \\
HunyuanAvatar~\cite{chen2025hunyuanavatar}   & 43.42 / 76.58  & 445.02 / 870.16   & \secondcell{0.87 / 0.91} & 4.92 / 2.71  & 10.01 / 9.34  & 14.40 \\
FantasyTalking~\cite{wang2025fantasytalking} & 43.14 / 144.71 & 483.11 / 1649.23  & 0.85 / 0.76  & 3.15 / 4.30  & 9.69 / 13.18  & 4.20 \\
OmniAvatar~\cite{gan2025omniavatar}          & 41.48 / 81.12  & 330.76 / 1430.79  & 0.89 / 0.86 & 5.58 / 4.21  & 9.01 / 10.82  & 6.21 \\
MultiTalk~\cite{kong2025multitalk}           & 44.31 / 106.04 & 419.19 / 1788.19  & 0.88 / 0.76 & 5.88 / 5.26   & 9.58/ 9.39    & 3.75 \\
StableAvatar~\cite{tu2025stableavatar}       & 38.94 / 58.80 & \secondcell{603.51 / 739.72}   & 0.88 / 0.90 & 3.42 / 2.35  & 10.99 / 11.38 & 7.40 \\
LiveAvatar (5 GPUs)~\cite{huang2025liveavatar}      & \secondcell{37.63 / 57.81} & 312.87 / 765.64 & 0.91 / 0.90 & \secondcell{6.28 / 5.46} & \bestcell{8.88 / 9.12}  & \secondcell{0.047} \\
WanS2V~\cite{gao2025wan}                    & 38.21 / 88.73 & 324.67 / 870.34 & 0.88 / 0.74 & 6.28 / 5.13 & 9.85 / 10.36 & 16.00 \\
AvatarForcing (Ours)                & \bestcell{37.54 / 56.22} & \bestcell{314.55 / 737.92} & \bestcell{0.91 / 0.91} & \bestcell{6.32 / 5.64} & \secondcell{8.79 / 9.26} & \bestcell{0.034} \\
\bottomrule
\end{tabular}
}
\end{table*}

\subsection{Distribution-Matching Post-Training}

To adapt the one-step generator to the distribution induced by rolling-window inference, we use Distribution Matching Distillation (DMD) with a frozen teacher score function $s_{\text{teach}}$ and a trainable critic $s_{\phi}$. Given a student prediction $\widehat{x}_0$ from rolling-window simulation, we sample a timestep $t$, construct $x_t$ by adding noise, estimate a KL gradient $g(x_t,t)$, and update the generator via
\begin{equation}
\mathcal{L}_{\text{DMD}}
=
\frac{1}{2}\left\Vert \widehat{x}_0 - \mathrm{stopgrad}\big(\widehat{x}_0 - g(x_t,t)\big)\right\Vert_2^2.
\end{equation}

\paragraph{Two-stage distillation with offline ODE backfill.}
The first stage records teacher trajectories via offline ODE backfill and performs audio-conditioned ODE regression, training a one-step predictor from intermediate noise levels to the clean endpoint. In our implementation, this ODE stage runs for 4{,}800 steps and provides a crucial initialization for one-step denoising across heterogeneous noise levels. The second stage applies DMD post-training on student rollouts from the same rolling-window inference procedure, including heterogeneous noise and KV caching, with lightweight mixed-window backpropagation. This DMD stage runs for 2{,}500 steps and corrects noise-level-dependent sharpness under heterogeneous-noise windows, avoiding multi-step teacher sampling in the loop while aligning training with streaming inference.

%% file: sec/4_experiment.tex
\section{Experiments}
\label{sec:exp}

\subsection{Setup and Long-Form Benchmark}
Training uses AVSpeech~\cite{avspeech} and an extended EMO corpus. A pretrained bidirectional audio--video diffusion teacher, initialized from WanS2V~\cite{gao2025wan}, provides distribution-matching targets. The training set contains approximately 500 hours of paired audio--video data spanning diverse speakers, emotional expressions, and recording conditions.

To evaluate streaming generation, we construct a long-form benchmark with nearly 400 videos ranging from 40 seconds to 2 minutes. We also report results on CelebV-HQ~\cite{zhu2022celebv}, a standard high-resolution benchmark.

We report fidelity (FID~\cite{fid}, FVD~\cite{fvd}), identity preservation (CSIM~\cite{javaheri2017convex}), and audio--visual alignment (Sync-C/Sync-D~\cite{syncsynd}). For long-form stability, we evaluate color drift ($\Delta E_{2000}$) and flicker (Adj-LPIPS on adjacent frames) on aligned face crops.

The system uses a pretrained 1.3B-parameter video diffusion backbone~\cite{wan2025basemodel} and integrates AvatarForcing $\mathcal{B}_{L,N}$ with dual-anchor KV caching under bounded attention. We generate 16{,}000 teacher ODE trajectory pairs for distribution-matching initialization and post-training (Sec.~\ref{sec:method}). The two-stage distillation runs 4{,}800 steps of ODE regression pre-training followed by 2{,}500 steps of DMD post-training. Videos are synthesized at 25 FPS with VAE latent spatial resolution $832 \times 480$. Unless stated otherwise, we use $B=4$ frames per block and $\mathcal{B}_{4,1}$ (window length $L=4$, one pass $N=1$), which yields a favorable quality--latency trade-off (Sec.~\ref{sec:exp_lvsn}). For audio conditioning, raw waveforms are encoded by a streaming speech encoder (Wav2Vec~\cite{baevski2020wav2vec}) into per-frame embeddings. The embeddings are window-aligned and injected via lightweight additive modulation with anchor-audio zeroing (Sec.~\ref{sec:audio}).

\begin{figure}[t]
	  \centering
	  \includegraphics[width=\linewidth]{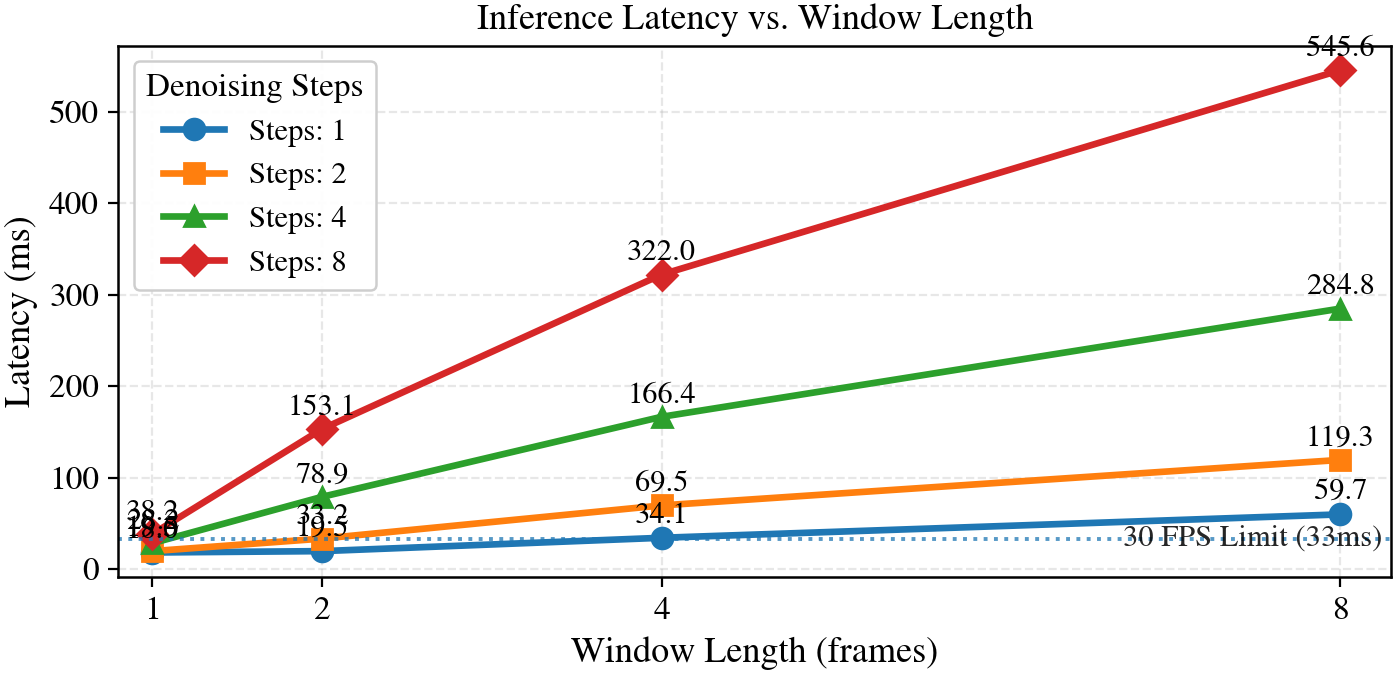}
	  \caption{\textbf{Latency vs.\ window length.} Inference latency increases with both window length and the number of denoising steps. Window length is shown in frames; with block size $B=4$, a window of $L$ blocks spans $BL$ frames.}
	  \label{fig:alb-latency}
\end{figure}

\subsection{Comparisons to State-of-the-Art}
We compare against recent audio-driven avatar models, including EchoMimic~\cite{chen2025echomimic}, Hallo3~\cite{cui2025hallo3}, HunyuanAvatar~\cite{chen2025hunyuanavatar}, FantasyTalking~\cite{wang2025fantasytalking}, OmniAvatar~\cite{gan2025omniavatar}, MultiTalk~\cite{kong2025multitalk}, and StableAvatar~\cite{tu2025stableavatar}. We additionally include LiveAvatar~\cite{huang2025liveavatar} and WanS2V~\cite{gao2025wan} as streaming-oriented and large-scale diffusion baselines. We follow the official implementations and recommended settings for all baselines.

Table~\ref{tab:main_quantitative} shows that AvatarForcing achieves the strongest performance on long-form videos, whereas other methods degrade over long rollouts with identity drift, color drift, and weaker lip synchronization (Fig.~\ref{fig:exp}). At 25 FPS and $832 \times 480$ latent resolution, the 1.3B-parameter student runs at 34\,ms/frame ($\approx 0.034\,s/frame$). Latency (s/frame) is measured as steady-state end-to-end wall-clock time per output frame, including streaming audio encoding, one DiT forward pass, KV cache update, and VAE decoding, while excluding model initialization, data loading, disk I/O, and video post-processing. All AvatarForcing timings are measured with batch size 1 on a single GPU using bfloat16 inference. We distinguish steady-state latency (s/frame) from end-to-end audio-to-visual delay; Tab.~\ref{tab:delay_protocol} summarizes the protocol and default values.

\begin{table}
\centering
\caption{\textbf{Steady-state latency vs.\ end-to-end delay protocol.} Default values use $L=4$, $B=4$, and 25 FPS.}
\label{tab:delay_protocol}
\small
\begin{tabular}{l p{0.45\linewidth} c}
\toprule
Quantity & Definition & Default \\
\midrule
Steady-state latency & Per-frame wall-clock time in steady state (audio encoder + DiT + KV update + VAE); excludes initialization, data loading, disk I/O, and post-processing. & $0.034$\,s \\
Audio look-ahead & Future audio required by bounded local-future denoising: $(L-1)B$ frames, $T_{\mathrm{LA}}=(L-1)B/25$. & $0.48$\,s \\
End-to-end delay & Measured wall-clock delay between receiving the audio embedding for frame $t$ and emitting the aligned video frame $t$ in a real-time streaming simulation (includes look-ahead buffering and steady-state compute). & $0.51$\,s \\
Emission unit & One clean block per step ($B$ frames); all timings are normalized per frame. & $B=4$ \\
\bottomrule
\end{tabular}
\end{table}

We provide additional qualitative comparisons, user studies, and failure cases.

\subsection{Ablation Studies}
Ablations isolate (i) how compute is allocated between bounded look-ahead ($L$) and per-step refinement ($N$) under a fixed streaming pipeline and (ii) how anchoring and distillation affect long-horizon stability under a fixed $\mathcal{B}_{L,N}$.

\subsubsection{Block Length \texorpdfstring{$L$}{L} vs.\ Denoising Steps \texorpdfstring{$N$}{N} in \texorpdfstring{$\mathcal{B}_{L,N}$}{B\_\{L,N\}}}
\label{sec:exp_lvsn}
\begin{figure}[t]
  \centering
  \includegraphics[width=\linewidth,height=0.23\linewidth]{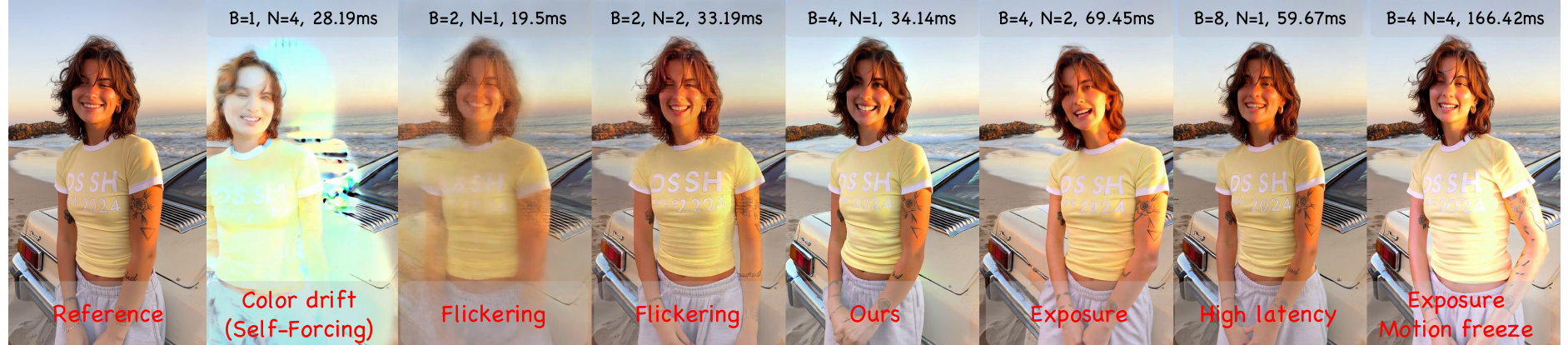}
  \caption{\textbf{Window length $L$ vs.\ denoising steps $N$.} Ablation on window lengths $L$ and denoising steps $N$ for $\mathcal{B}_{L,N}$.}
  \label{fig:alb-window}
  \vspace{-10pt}
\end{figure}

AvatarForcing trades denoising depth for bounded look-ahead by jointly denoising an $L$-block window with $N$ passes. In this sweep, we fix the dual-anchor KV cache, audio alignment, and block size $B$; varying $L$ only determines whether denoising can condition on future (noisy) blocks. Setting $L{=}1$ removes local-future look-ahead and yields a strictly causal block-level update, where the model relies on cached history and the two anchors. We sweep $(L,N)$ and report stability and flicker in Fig.~\ref{fig:alb-window} and Tab.~\ref{tab:alb-window}, with measured latency in Fig.~\ref{fig:alb-latency}. Under a matched budget of $L\cdot N$ (and similar latency), increasing $L$ often improves long-horizon stability more than increasing $N$ (e.g., $\mathcal{B}_{4,1}$ versus $\mathcal{B}_{2,2}$), which suggests that the gains primarily arise from bounded look-ahead rather than from repeated refinement under a short context. Overly large $L$ (or $L\cdot N$) increases latency and can over-smooth motion. Unless stated otherwise, we use $\mathcal{B}_{4,1}$ as the default.

\definecolor{lat1}{HTML}{FFFFFF}
\definecolor{lat2}{HTML}{F7FBFF}
\definecolor{lat4}{HTML}{DEEBF7}
\definecolor{lat8}{HTML}{C6DBEF}
\definecolor{lat16}{HTML}{9ECAE1}
\definecolor{lat32}{HTML}{6BAED6}
\definecolor{lat64}{HTML}{4292C6}
\newcommand{\prodcell}[3]{
  \begingroup
  \count0=\numexpr#1*#2\relax
  \ifnum\count0=1 \cellcolor{lat1}\fi
  \ifnum\count0=2 \cellcolor{lat2}\fi
  \ifnum\count0=4 \cellcolor{lat4}\fi
  \ifnum\count0=8 \cellcolor{lat8}\fi
  \ifnum\count0=16 \cellcolor{lat16}\fi
  \ifnum\count0=32 \cellcolor{lat32}\fi
  \ifnum\count0=64 \cellcolor{lat64}\fi
  #3%
  \endgroup
}

\begin{table}
\centering
\captionsetup{justification=raggedright,singlelinecheck=false}
\caption{\textbf{Grid ablation of window length $L$ and denoising steps $N$.} Entries report CSIM $\uparrow$ / Adj-LPIPS $\downarrow$ / latency (ms) $\downarrow$; Adj-LPIPS is reported in $10^{-2}$. Cells with the same compute budget $L\cdot N$ share the same background color (blue gradient).}
\label{tab:alb-window}
\resizebox{\linewidth}{!}{
\begin{tabular}{c|cccc}
\toprule
Denoising step $N$ $\backslash$ block length $L$ & 1 & 2 & 4 & 8 \\
\midrule
1  & \prodcell{1}{1}{0.07/0.10/31.4} & \prodcell{2}{1}{0.73/3.23/19.5} & \prodcell{4}{1}{0.90/1.06/34.14} & \prodcell{8}{1}{0.91/0.84/59.67} \\
2  & \prodcell{1}{2}{0.52/8.12/18.66} & \prodcell{2}{2}{0.83/2.25/33.19} & \prodcell{4}{2}{0.80/1.94/69.45} & \prodcell{8}{2}{0.75/0.72/119.32} \\
4  & \prodcell{1}{4}{0.32/8.12/28.19} & \prodcell{2}{4}{0.87/2.23/78.94} & \prodcell{4}{4}{0.90/0.75/166.42} & \prodcell{8}{4}{0.82/0.83/284.79} \\
8  & \prodcell{1}{8}{0.74/2.24/38.2} & \prodcell{2}{8}{0.91/1.10/153.07} & \prodcell{4}{8}{0.81/1.20/321.98} & \prodcell{8}{8}{0.84/0.87/545.62} \\
\bottomrule
\end{tabular}
}
\end{table}

\subsubsection{Teacher/Student and Dual-Anchor KV Ablations}
\begin{figure}[t]
  \centering
  \vspace{-10pt}
  \includegraphics[width=\linewidth,height=0.33\linewidth]{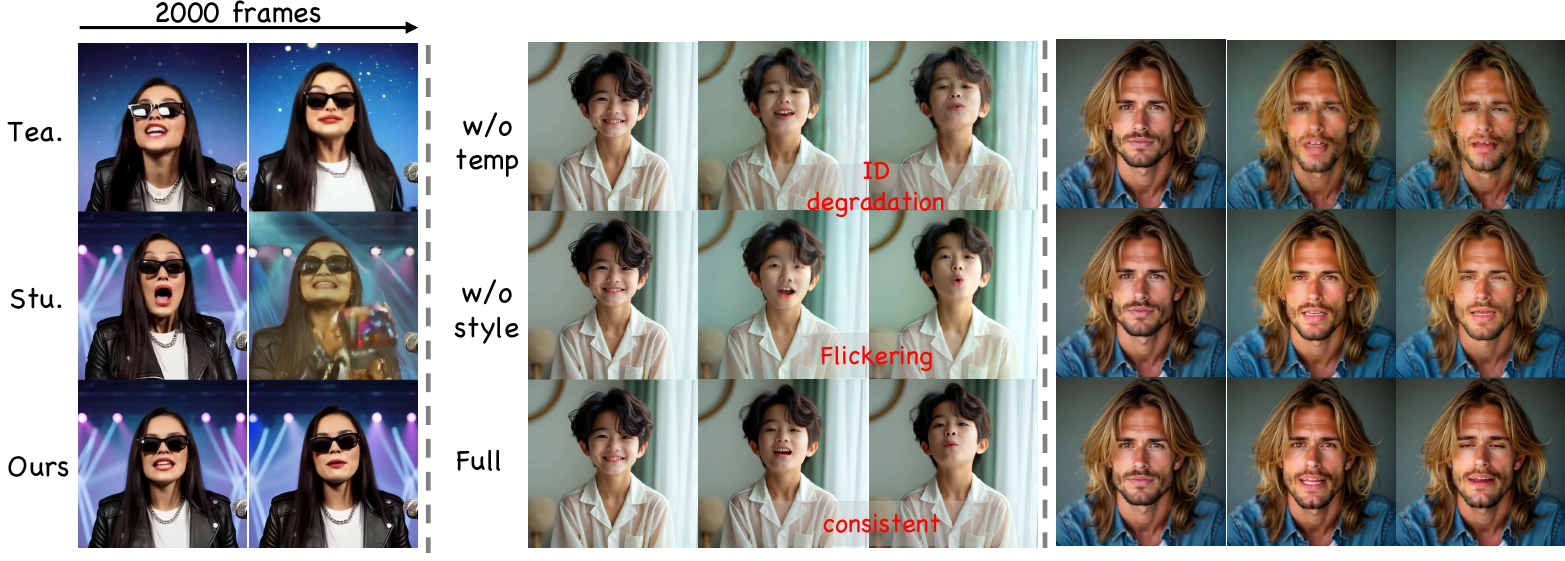}
  \caption{\textbf{Teacher/student and anchor ablation.} Left: Teacher/Student/Ours. Right: removing either anchor.}
  \label{fig:alb-teacher}
  \vspace{-10pt}
\end{figure}

This study isolates the effects of the distillation backbone and the dual-anchor KV cache under long-form streaming. Tab.~\ref{tab:alb-teacher} (left) compares the offline bidirectional teacher, a distilled student, and AvatarForcing. Latency is reported in seconds per output frame (s/frame). The teacher provides strong supervision but is not suitable for real-time deployment due to high latency (16\,s/frame). The distilled student is faster but less stable, whereas AvatarForcing remains real-time (34\,ms/frame) and improves CSIM while reducing color drift ($\Delta E_{2000}$) and flicker (Adj-LPIPS). Tab.~\ref{tab:alb-teacher} (right) confirms that both anchors are necessary: removing either anchor increases drift and flicker. Fig.~\ref{fig:alb-teacher} visualizes these failure modes under long rollouts.

\begin{table*}
\centering
\captionsetup{justification=raggedright,singlelinecheck=false}
\caption{\textbf{Ablation on distillation backbone and anchors.} Left: Teacher/Stud- ent/ Ours. Right: removing either anchor. Latency is reported in s/frame. Long-form stability metrics are reported as Total / Last over 1-minute rollouts.}
\label{tab:alb-teacher}
\begin{minipage}[t]{0.47\textwidth}
\centering
\resizebox{\linewidth}{!}{
\begin{tabular}{lcccc}
\toprule
Model & Latency (s/frame)$\downarrow$ & CSIM$\uparrow$ & $\mathbf{\Delta E_{2000}}\downarrow$ & Adj-LPIPS$\downarrow$ \\
\midrule
Teacher & 16 &        0.88/0.84 & 18.51/24.18 & 0.18/0.28 \\
Student & 4.8 & 0.86/0.84 & 17.77/25.86 & 0.09/0.15 \\
AvatarForcing & 0.034 & 0.90/0.86 & 8.71/11.39 & 0.0106/0.0189 \\
\bottomrule
\end{tabular}
}
\end{minipage}\hfill
\begin{minipage}[t]{0.51\textwidth}
\centering
\resizebox{\linewidth}{!}{
\begin{tabular}{lcccc}
\toprule
Variant & Latency (s/frame)$\downarrow$ & CSIM$\uparrow$ & $\mathbf{\Delta E_{2000}}\downarrow$ & Adj-LPIPS$\downarrow$ \\
\midrule
w/o style anchor & 0.034 & 0.72/0.63 & 9.63/12.68 & 0.0196/0.0209 \\
w/o temporal anchor & 0.031 & 0.86/0.63 & 10.70/14.29 & 0.0109/0.0209 \\
Full & 0.034 & 0.90/0.86 & 8.71/11.39 & 0.0106/0.0189 \\
\bottomrule
\end{tabular}
}
\end{minipage}
\end{table*}

\begin{table*}
\centering
\caption{\textbf{Anchor-audio zero padding and RoPE re-indexing.} Ablation on anchor-audio zero padding and RoPE re-indexing. Metrics follow Tab.~\ref{tab:onestep_baselines}.}
\label{tab:alb_zeropad_rope}
\resizebox{\linewidth}{!}{
\begin{tabular}{lccccccc}
\toprule
Variant & FID$\downarrow$ & FVD$\downarrow$ & CSIM$\uparrow$ & Sync-C$\uparrow$ & Sync-D$\downarrow$ & $\mathbf{\Delta E_{2000}}\downarrow$ & Adj-LPIPS$\downarrow$ \\
\midrule
w/o anchor-audio zero padding & 58.53 & 796.35 & 0.89/0.84 & 3.43 & 11.23 & 8.91/13.87 & 0.0110/0.0191 \\
w/o RoPE re-indexing & 72.43 & 1382.63 & 0.67/0.49 & 2.23 & 19.23  & 13.56/33.86 & 0.018/0.032 \\
Full & 56.22 & 737.92 & 0.90/0.86 & 5.64 & 9.26  & 8.71/11.39 & 0.0106/0.0189 \\
\bottomrule
\end{tabular}
}
\end{table*}

\begin{table*}[!t]
\centering
\caption{\textbf{One-step causal baselines on long-form videos.} Baselines are adapted to one-step inference. Note that Causal Forcing is the cascade of Causal ODE initialization and DMD; here we only use Causal ODE as a baseline rather than the full method. FID/FVD/Sync are reported on long-form videos (Sec.~\ref{sec:exp}); long-horizon stability metrics are reported as Total / Last over 1-minute rollouts.}
\label{tab:onestep_baselines}
\resizebox{\linewidth}{!}{
\begin{tabular}{lcccccccc}
\toprule
Method & FID$\downarrow$ & FVD$\downarrow$ & CSIM$\uparrow$ & Sync-C$\uparrow$ & Sync-D$\downarrow$ & Latency (s/frame)$\downarrow$ & $\mathbf{\Delta E_{2000}}\downarrow$ & Adj-LPIPS$\downarrow$ \\
\midrule
Self-Forcing (1-step)~\cite{huang2025selfforcing} & 88.52 & 1127.43 & 0.40/0.07 & 1.34 & 19.53 & 0.031 & 27.91/40.63 & 0.05/0.10 \\
Causal ODE~\cite{zhu2026causalforcing} & 102.63 & 1382.63 & 0.58/0.32 & 0.12 & 24.21  & 0.029 &  16.73/35.92 & 0.08/0.12 \\
AvatarForcing (Ours) & 56.22 & 737.92 & 0.90/0.86 & 5.64 & 9.26 & 0.034 & 8.71/11.39 & 0.0106/0.0189 \\
\bottomrule
\end{tabular}
}
\end{table*}

\subsubsection{One-Step Streaming Baselines}
\begin{figure}[!t]
	  \centering
	  \includegraphics[width=\linewidth]{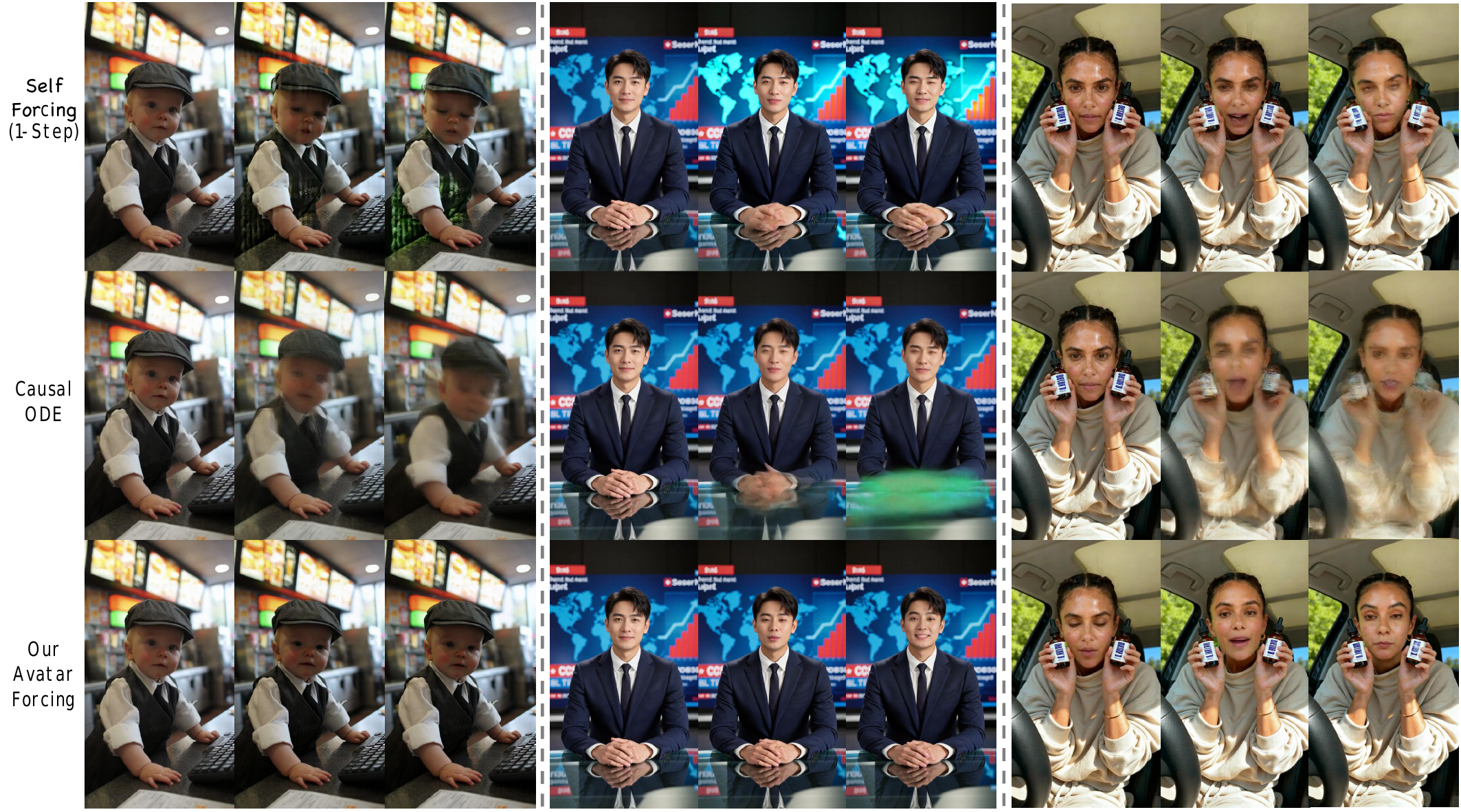}
	  \caption{\textbf{Qualitative comparison to one-step causal baselines.} Self-Forcing (1-step) denotes our one-step DMD baseline and shows noticeable color drift, while Causal ODE denotes our one-step ODE baseline and produces blurred frames with noise-level-dependent sharpness. AvatarForcing better preserves appearance consistency and sharp motion under long rollouts, and the ODE-only blur motivates the two-stage design (ODE initialization followed by DMD refinement).}
	  \label{fig:forcing}
    \vspace{-10pt}
\end{figure}

We further compare against strictly causal one-step baselines by adapting forcing-style pipelines to the same one-step distillation setting. Self-Forcing (1-step) adapts the Self-Forcing pipeline~\cite{huang2025selfforcing} by replacing multi-step denoising with a single denoising update for each predicted block. Causal ODE is a one-step causal baseline derived from Causal Forcing~\cite{zhu2026causalforcing} by adopting the causal ODE distillation recipe while maintaining one-step inference. Note that Causal Forcing is the cascade of Causal ODE initialization and DMD; here we only use Causal ODE as a baseline rather than the full method. Tab.~\ref{tab:onestep_baselines} reports quantitative results, and Fig.~\ref{fig:forcing} shows qualitative comparisons under long rollouts.

\subsubsection{Additional Ablations}
\begin{figure}[!t]
  \centering
  \includegraphics[width=\linewidth]{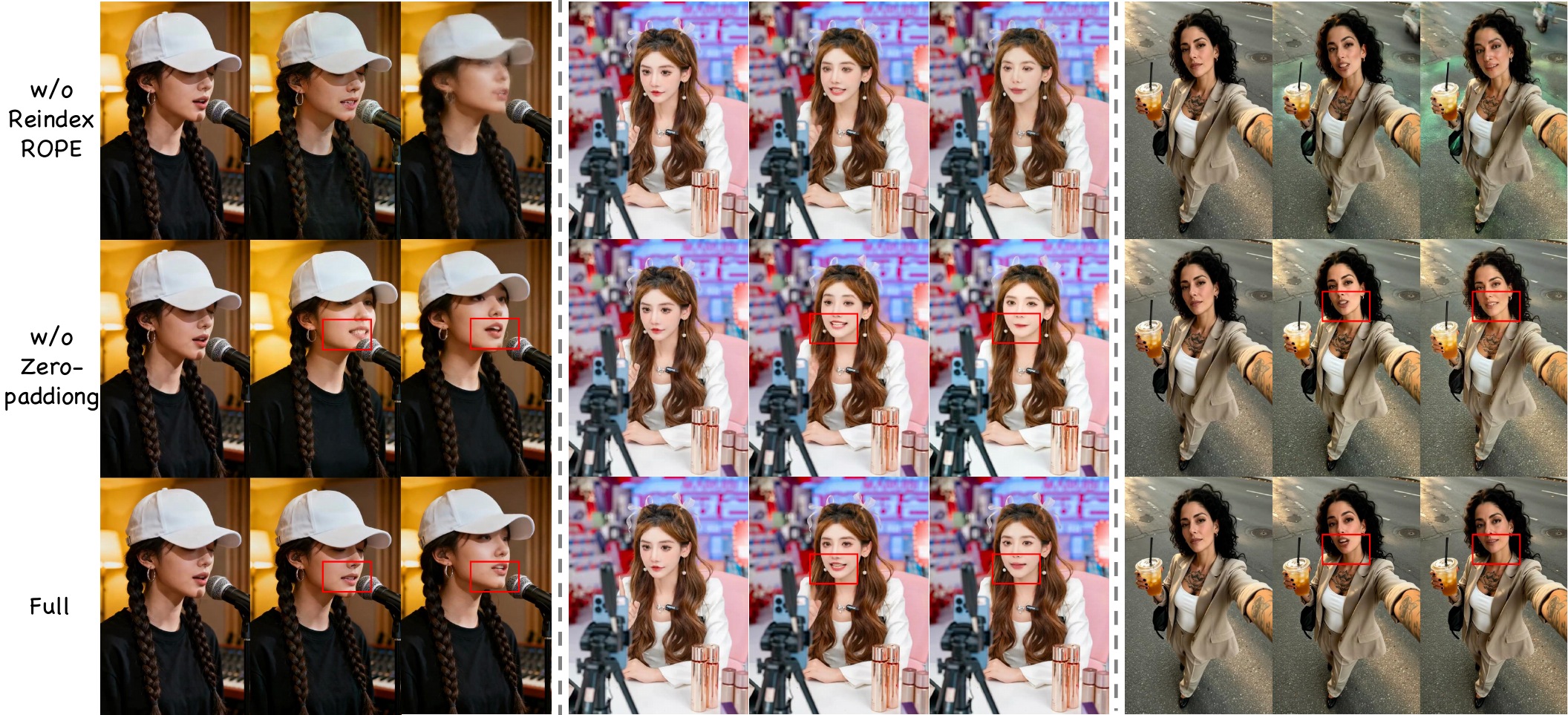}
  \caption{\textbf{Qualitative ablation on anchor-audio zero padding and RoPE re-indexing.} Red boxes highlight mouth regions. Without anchor-audio zero padding, misaligned anchor audio contaminates the anchor pathway and induces mouth jitter and artifacts. Without RoPE re-indexing, the style anchor becomes phase-misaligned as the stream grows, leading to gradual appearance and color drift. The full model avoids both failure modes, consistent with Tab.~\ref{tab:alb_zeropad_rope}.}
  \label{fig:alb_zeropad_rope}
\end{figure}

We ablate two design choices that stabilize the style anchor under unbounded streaming: anchor-audio zero padding and RoPE re-indexing. Without anchor-audio zero padding, the style anchor receives nonzero speech features from the current window, which injects speech dynamics into the anchor pathway and produces spurious mouth motion and local distortions (red boxes in Fig.~\ref{fig:alb_zeropad_rope}). Without RoPE re-indexing, the style anchor becomes increasingly distant in absolute time as the window slides, weakening identity anchoring and causing blurred frames and appearance drift over long rollouts. Tab.~\ref{tab:alb_zeropad_rope} confirms these effects: removing either component degrades perceptual quality and long-form stability, with RoPE re-indexing being particularly important for preventing drift and flicker. We also ablate teacher CFG scale and timestep-shift scheduling, and report a detailed latency breakdown and long-horizon failure cases.

%% file: sec/5_conclusion.tex
\vspace{-0.5 em}
\section{Conclusion}
\label{sec:con}
\vspace{-0.5 em}

We present AvatarForcing, a one-step streaming diffusion framework for real-time, long-form talking-avatar synthesis that maintains temporal consistency and accurate audio synchronization. The method uses the $\mathcal{B}_{L,N}$ sliding-window joint denoising mechanism, so each block is refined multiple times before emission. To preserve coherence beyond the bounded window, AvatarForcing uses a dual-anchor KV cache with a RoPE re-indexed style anchor for identity and a temporal anchor for smooth transitions. Per-frame audio is aligned to the active window, and anchor-audio zeroing prevents cross-window interference. A two-stage distillation procedure transfers supervision from a global bidirectional teacher through distribution matching, which enables 34\,ms/frame inference. Larger windows can reduce drift but increase latency, and very long generation can still degrade when the bounded context becomes insufficient. Future work will study stronger long-range memory and improved motion preservation.






%% file: main.bbl
\begin{thebibliography}{10}
\providecommand{\url}[1]{#1}
\csname url@samestyle\endcsname
\providecommand{\newblock}{\relax}
\providecommand{\bibinfo}[2]{#2}
\providecommand{\BIBentrySTDinterwordspacing}{\spaceskip=0pt\relax}
\providecommand{\BIBentryALTinterwordstretchfactor}{4}
\providecommand{\BIBentryALTinterwordspacing}{\spaceskip=\fontdimen2\font plus
\BIBentryALTinterwordstretchfactor\fontdimen3\font minus \fontdimen4\font\relax}
\providecommand{\BIBforeignlanguage}[2]{{%
\expandafter\ifx\csname l@#1\endcsname\relax
\typeout{** WARNING: IEEEtran.bst: No hyphenation pattern has been}%
\typeout{** loaded for the language `#1'. Using the pattern for}%
\typeout{** the default language instead.}%
\else
\language=\csname l@#1\endcsname
\fi
#2}}
\providecommand{\BIBdecl}{\relax}
\BIBdecl

\bibitem{tian2025emo2}
L.~Tian, S.~Hu, Q.~Wang, B.~Zhang, and L.~Bo, ``Emo2: End-effector guided audio-driven avatar video generation,'' \emph{arXiv preprint arXiv:2501.10687}, 2025.

\bibitem{ding2025kling}
Y.~Ding, J.~Liu, W.~Zhang, Z.~Wang, W.~Hu, L.~Cui, M.~Lao, Y.~Shao, H.~Liu, X.~Li \emph{et~al.}, ``Kling-avatar: Grounding multimodal instructions for cascaded long-duration avatar animation synthesis,'' \emph{arXiv preprint arXiv:2509.09595}, 2025.

\bibitem{gao2025wan}
X.~Gao, L.~Hu, S.~Hu, M.~Huang, C.~Ji, D.~Meng, J.~Qi, P.~Qiao, Z.~Shen, Y.~Song \emph{et~al.}, ``Wan-s2v: Audio-driven cinematic video generation,'' \emph{arXiv preprint arXiv:2508.18621}, 2025.

\bibitem{wan2025basemodel}
T.~Wan, A.~Wang, B.~Ai, B.~Wen, C.~Mao, C.-W. Xie, D.~Chen, F.~Yu, H.~Zhao, J.~Yang, J.~Zeng, J.~Wang, J.~Zhang, J.~Zhou, J.~Wang, J.~Chen, K.~Zhu, K.~Zhao, K.~Yan, L.~Huang, M.~Feng, N.~Zhang, P.~Li, P.~Wu, R.~Chu, R.~Feng, S.~Zhang, S.~Sun, T.~Fang, T.~Wang, T.~Gui, T.~Weng, T.~Shen, W.~Lin, W.~Wang, W.~Wang, W.~Zhou, W.~Wang, W.~Shen, W.~Yu, X.~Shi, X.~Huang, X.~Xu, Y.~Kou, Y.~Lv, Y.~Li, Y.~Liu, Y.~Wang, Y.~Zhang, Y.~Huang, Y.~Li, Y.~Wu, Y.~Liu, Y.~Pan, Y.~Zheng, Y.~Hong, Y.~Shi, Y.~Feng, Z.~Jiang, Z.~Han, Z.-F. Wu, and Z.~Liu, ``Wan: Open and advanced large-scale video generative models,'' \emph{arXiv preprint arXiv:2503.20314}, 2025.

\bibitem{zhang2025waver}
Y.~Zhang, H.~Yang, Y.~Zhang, Y.~Hu, F.~Zhu, C.~Lin, X.~Mei, Y.~Jiang, B.~Peng, and Z.~Yuan, ``Waver: Wave your way to lifelike video generation,'' \emph{arXiv preprint arXiv:2508.15761}, 2025.

\bibitem{gao2025seedance}
Y.~Gao, H.~Guo, T.~Hoang, W.~Huang, L.~Jiang, F.~Kong, H.~Li, J.~Li, L.~Li, X.~Li \emph{et~al.}, ``Seedance 1.0: Exploring the boundaries of video generation models,'' \emph{arXiv preprint arXiv:2506.09113}, 2025.

\bibitem{liu2024anitalker}
T.~Liu, F.~Chen, S.~Fan, C.~Du, Q.~Chen, X.~Chen, and K.~Yu, ``Anitalker: animate vivid and diverse talking faces through identity-decoupled facial motion encoding,'' in \emph{Proceedings of the 32nd ACM International Conference on Multimedia}, 2024, pp. 6696--6705.

\bibitem{zhang2025xactro}
C.~Zhang, Z.~Li, H.~Xu, Y.~Xie, X.~Zhao, T.~Gu, G.~Song, X.~Chen, C.~Liang, J.~Jiang \emph{et~al.}, ``X-actor: Emotional and expressive long-range portrait acting from audio,'' \emph{arXiv preprint arXiv:2508.02944}, 2025.

\bibitem{gu2025diffusion}
Z.~Gu, R.~Yan, J.~Lu, P.~Li, Z.~Dou, C.~Si, Z.~Dong, Q.~Liu, C.~Lin, Z.~Liu \emph{et~al.}, ``Diffusion as shader: 3d-aware video diffusion for versatile video generation control,'' in \emph{Proceedings of the Special Interest Group on Computer Graphics and Interactive Techniques Conference Conference Papers}, 2025, pp. 1--12.

\bibitem{wang2024motionctrl}
Z.~Wang, Z.~Yuan, X.~Wang, Y.~Li, T.~Chen, M.~Xia, P.~Luo, and Y.~Shan, ``Motionctrl: A unified and flexible motion controller for video generation,'' in \emph{ACM SIGGRAPH 2024 Conference Papers}, 2024, pp. 1--11.

\bibitem{chen2025midas}
M.~Chen, L.~Cui, W.~Zhang, H.~Zhang, Y.~Zhou, X.~Li, S.~Tang, J.~Liu, B.~Liao, H.~Chen \emph{et~al.}, ``Midas: Multimodal interactive digital-human synthesis via real-time autoregressive video generation,'' \emph{arXiv preprint arXiv:2508.19320}, 2025.

\bibitem{casvid}
T.~Yin, Q.~Zhang, R.~Zhang, W.~T. Freeman, F.~Durand, E.~Shechtman, and X.~Huang, ``From slow bidirectional to fast autoregressive video diffusion models,'' in \emph{Proceedings of the Computer Vision and Pattern Recognition Conference}, 2025, pp. 22\,963--22\,974.

\bibitem{teng2025magi}
H.~Teng, H.~Jia, L.~Sun, L.~Li, M.~Li, M.~Tang, S.~Han, T.~Zhang, W.~Zhang, W.~Luo \emph{et~al.}, ``Magi-1: Autoregressive video generation at scale,'' \emph{arXiv preprint arXiv:2505.13211}, 2025.

\bibitem{deng2024autoregressive}
H.~Deng, T.~Pan, H.~Diao, Z.~Luo, Y.~Cui, H.~Lu, S.~Shan, Y.~Qi, and X.~Wang, ``Autoregressive video generation without vector quantization,'' \emph{arXiv preprint arXiv:2412.14169}, 2024.

\bibitem{low2025talkingmachines}
C.~Low and W.~Wang, ``Talkingmachines: Real-time audio-driven facetime-style video via autoregressive diffusion models,'' \emph{arXiv preprint arXiv:2506.03099}, 2025.

\bibitem{yang2025longlive}
S.~Yang, W.~Huang, R.~Chu, Y.~Xiao, Y.~Zhao, X.~Wang, M.~Li, E.~Xie, Y.~Chen, Y.~Lu \emph{et~al.}, ``Longlive: Real-time interactive long video generation,'' \emph{arXiv preprint arXiv:2509.22622}, 2025.

\bibitem{huang2025selfforcing}
X.~Huang, Z.~Li, G.~He, M.~Zhou, and E.~Shechtman, ``Self forcing: Bridging the train-test gap in autoregressive video diffusion,'' \emph{arXiv preprint arXiv:2506.08009}, 2025.

\bibitem{cui2025selfforcingpp}
J.~Cui, J.~Wu, M.~Li, T.~Yang, X.~Li, R.~Wang, A.~Bai, Y.~Ban, and C.-J. Hsieh, ``Self-forcing++: Towards minute-scale high-quality video generation,'' \emph{arXiv preprint arXiv:2510.02283}, 2025.

\bibitem{chen2026contextforcing}
S.~Chen, C.~Wei, S.~Sun, P.~Nie, K.~Zhou, G.~Zhang, M.-H. Yang, and W.~Chen, ``Context forcing: Consistent autoregressive video generation with long context,'' \emph{arXiv preprint arXiv:2602.06028}, 2026.

\bibitem{wang2025fantasytalking}
M.~Wang, Q.~Wang, F.~Jiang, Y.~Fan, Y.~Zhang, Y.~Qi, K.~Zhao, and M.~Xu, ``Fantasytalking: Realistic talking portrait generation via coherent motion synthesis,'' \emph{arXiv preprint arXiv:2504.04842}, 2025.

\bibitem{hello3}
H.~Kuntz, K.~Kimbel, M.~Bucher, E.~VanDyke, and L.~J. Van~Scoy, ``Facilitating advance care planning for patients with cancer via a conversation game,'' \emph{Journal of Pain and Symptom Management}, vol.~69, no.~5, pp. e604--e605, 2025.

\bibitem{chen2025hunyuanvideo}
Y.~Chen, S.~Liang, Z.~Zhou, Z.~Huang, Y.~Ma, J.~Tang, Q.~Lin, Y.~Zhou, and Q.~Lu, ``Hunyuanvideo-avatar: High-fidelity audio-driven human animation for multiple characters,'' \emph{arXiv preprint arXiv:2505.20156}, 2025.

\bibitem{tian2024emo}
L.~Tian, Q.~Wang, B.~Zhang, and L.~Bo, ``Emo: Emote portrait alive generating expressive portrait videos with audio2video diffusion model under weak conditions,'' in \emph{European Conference on Computer Vision}.\hskip 1em plus 0.5em minus 0.4em\relax Springer, 2024, pp. 244--260.

\bibitem{lin2025omnihuman1}
G.~Lin, J.~Jiang, J.~Yang, Z.~Zheng, C.~Liang, Y.~Zhang, and J.~Liu, ``Omnihuman-1: Rethinking the scaling-up of one-stage conditioned human animation models,'' in \emph{Proceedings of the IEEE/CVF International Conference on Computer Vision}, 2025, pp. 13\,847--13\,858.

\bibitem{jiang2025omnihuman1.5}
J.~Jiang, W.~Zeng, Z.~Zheng, J.~Yang, C.~Liang, W.~Liao, H.~Liang, Y.~Zhang, and M.~Gao, ``Omnihuman-1.5: Instilling an active mind in avatars via cognitive simulation,'' \emph{arXiv preprint arXiv:2508.19209}, 2025.

\bibitem{sun2025streamavatar}
Z.~Sun, Z.~Peng, Y.~Ma, Y.~Chen, Z.~Zhou, Z.~Zhou, G.~Zhang, Y.~Zhang, Y.~Zhou, Q.~Lu, and Y.-J. Liu, ``Streamavatar: Streaming diffusion models for real-time interactive human avatars,'' \emph{arXiv preprint arXiv:2512.22065}, 2025.

\bibitem{sun2025ardiffusion}
M.~Sun, W.~Wang, G.~Li, J.~Liu, J.~Sun, W.~Feng, S.~Lao, S.~Zhou, Q.~He, and J.~Liu, ``Ar-diffusion: Asynchronous video generation with auto-regressive diffusion,'' in \emph{Proceedings of the Computer Vision and Pattern Recognition Conference}, 2025, pp. 7364--7373.

\bibitem{cheng2025playing}
X.~Cheng, T.~He, J.~Xu, J.~Guo, D.~He, and J.~Bian, ``Playing with transformer at 30+ fps via next-frame diffusion,'' \emph{arXiv preprint arXiv:2506.01380}, 2025.

\bibitem{gao2024ca2}
K.~Gao, J.~Shi, H.~Zhang, C.~Wang, J.~Xiao, and L.~Chen, ``Ca2-vdm: Efficient autoregressive video diffusion model with causal generation and cache sharing,'' \emph{arXiv preprint arXiv:2411.16375}, 2024.

\bibitem{xu2024msc}
X.~Xu and M.~Cao, ``Msc: Multi-scale spatio-temporal causal attention for autoregressive video diffusion,'' \emph{arXiv preprint arXiv:2412.09828}, 2024.

\bibitem{chen2024diffusionforcing}
B.~Chen, D.~Mart{\'\i}~Mons{\'o}, Y.~Du, M.~Simchowitz, R.~Tedrake, and V.~Sitzmann, ``Diffusion forcing: Next-token prediction meets full-sequence diffusion,'' \emph{Advances in Neural Information Processing Systems}, vol.~37, pp. 24\,081--24\,125, 2024.

\bibitem{guo2025long}
Y.~Guo, C.~Yang, Z.~Yang, Z.~Ma, Z.~Lin, Z.~Yang, D.~Lin, and L.~Jiang, ``Long context tuning for video generation,'' \emph{arXiv preprint arXiv:2503.10589}, 2025.

\bibitem{zhu2026causalforcing}
H.~Zhu, M.~Zhao, G.~He, H.~Su, C.~Li, and J.~Zhu, ``Causal forcing: Autoregressive diffusion distillation done right for high-quality real-time interactive video generation,'' \emph{arXiv preprint arXiv:2602.02214}, 2026.

\bibitem{chen2024streaming}
F.~Chen, Z.~Yang, B.~Zhuang, and Q.~Wu, ``Streaming video diffusion: Online video editing with diffusion models,'' \emph{arXiv preprint arXiv:2405.19726}, 2024.

\bibitem{zheng2024memo}
L.~Zheng, Y.~Zhang, H.~Guo, J.~Pan, Z.~Tan, J.~Lu, C.~Tang, B.~An, and S.~Yan, ``Memo: Memory-guided diffusion for expressive talking video generation,'' \emph{arXiv preprint arXiv:2412.04448}, 2024.

\bibitem{fan2025syncdiff}
X.~Fan, H.~Gao, Z.~Chen, P.~Chang, M.~Han, and M.~Hasegawa-Johnson, ``Syncdiff: Diffusion-based talking head synthesis with bottlenecked temporal visual prior for improved synchronization,'' in \emph{2025 IEEE/CVF Winter Conference on Applications of Computer Vision (WACV)}.\hskip 1em plus 0.5em minus 0.4em\relax IEEE, 2025, pp. 4554--4563.

\bibitem{li2025ditto}
T.~Li, R.~Zheng, M.~Yang, J.~Chen, and M.~Yang, ``Ditto: Motion-space diffusion for controllable realtime talking head synthesis,'' in \emph{Proceedings of the 33rd ACM International Conference on Multimedia}, 2025, pp. 9704--9713.

\bibitem{kim2024fifodiffusion}
J.~Kim, J.~Kang, J.~Choi, and B.~Han, ``Fifo-diffusion: Generating infinite videos from text without training,'' \emph{Advances in Neural Information Processing Systems}, 2024.

\bibitem{kodaira2025streamdit}
A.~Kodaira, T.~Hou, J.~Hou, M.~Tomizuka, and Y.~Zhao, ``Streamdit: Real-time streaming text-to-video generation,'' \emph{arXiv preprint arXiv:2507.03745}, 2025.

\bibitem{huang2025liveavatar}
Y.~Huang, H.~Guo, F.~Wu, S.~Zhang, S.~Huang, Q.~Gan, L.~Liu, S.~Zhao, E.~Chen, J.~Liu, and S.~Hoi, ``Live avatar: Streaming real-time audio-driven avatar generation with infinite length,'' \emph{arXiv preprint arXiv:2512.04677}, 2025.

\bibitem{gu2025far}
Y.~Gu, W.~Mao, and M.~Z. Shou, ``Long-context autoregressive video modeling with next-frame prediction,'' \emph{arXiv preprint arXiv:2503.19325}, 2025.

\bibitem{chen2025echomimic}
Z.~Chen, J.~Cao, Z.~Chen, Y.~Li, and C.~Ma, ``Echomimic: Lifelike audio-driven portrait animations through editable landmark conditions,'' in \emph{Proceedings of the AAAI Conference on Artificial Intelligence}, vol.~39, no.~3, 2025, pp. 2403--2410.

\bibitem{cui2025hallo3}
J.~Cui, H.~Li, Y.~Zhan, H.~Shang, K.~Cheng, Y.~Ma, S.~Mu, H.~Zhou, J.~Wang, and S.~Zhu, ``Hallo3: Highly dynamic and realistic portrait image animation with video diffusion transformer,'' in \emph{Proceedings of the Computer Vision and Pattern Recognition Conference}, 2025, pp. 21\,086--21\,095.

\bibitem{chen2025hunyuanavatar}
Y.~Chen, S.~Liang, Z.~Zhou, Z.~Huang, Y.~Ma, J.~Tang, Q.~Lin, Y.~Zhou, and Q.~Lu, ``Hunyuanvideo-avatar: High-fidelity audio-driven human animation for multiple characters,'' \emph{arXiv preprint arXiv:2505.20156}, 2025.

\bibitem{gan2025omniavatar}
Q.~Gan, R.~Yang, J.~Zhu, S.~Xue, and S.~Hoi, ``Omniavatar: Efficient audio-driven avatar video generation with adaptive body animation,'' \emph{arXiv preprint arXiv:2506.18866}, 2025.

\bibitem{kong2025multitalk}
Z.~Kong, F.~Gao, Y.~Zhang, Z.~Kang, X.~Wei, X.~Cai, G.~Chen, and W.~Luo, ``Let them talk: Audio-driven multi-person conversational video generation,'' \emph{arXiv preprint arXiv:2505.22647}, 2025.

\bibitem{tu2025stableavatar}
S.~Tu, Y.~Pan, Y.~Huang, X.~Han, Z.~Xing, Q.~Dai, C.~Luo, Z.~Wu, and Y.-G. Jiang, ``Stableavatar: Infinite-length audio-driven avatar video generation,'' \emph{arXiv preprint arXiv:2508.08248}, 2025.

\bibitem{avspeech}
A.~Ephrat, I.~Mosseri, O.~Lang, T.~Dekel, K.~Wilson, A.~Hassidim, W.~T. Freeman, and M.~Rubinstein, ``Looking to listen at the cocktail party: A speaker-independent audio-visual model for speech separation,'' \emph{arXiv preprint arXiv:1804.03619}, 2018.

\bibitem{zhu2022celebv}
H.~Zhu, W.~Wu, W.~Zhu, L.~Jiang, S.~Tang, L.~Zhang, Z.~Liu, and C.~C. Loy, ``Celebv-hq: A large-scale video facial attributes dataset,'' in \emph{European conference on computer vision}.\hskip 1em plus 0.5em minus 0.4em\relax Springer, 2022, pp. 650--667.

\bibitem{fid}
M.~Heusel, H.~Ramsauer, T.~Unterthiner, B.~Nessler, and S.~Hochreiter, ``Gans trained by a two time-scale update rule converge to a local nash equilibrium,'' \emph{Advances in neural information processing systems}, vol.~30, 2017.

\bibitem{fvd}
T.~Unterthiner, S.~Van~Steenkiste, K.~Kurach, R.~Marinier, M.~Michalski, and S.~Gelly, ``Towards accurate generative models of video: A new metric \& challenges,'' \emph{arXiv preprint arXiv:1812.01717}, 2018.

\bibitem{javaheri2017convex}
A.~Javaheri, H.~Zayyani, and F.~Marvasti, ``A convex similarity index for sparse recovery of missing image samples,'' \emph{arXiv preprint arXiv:1701.07422}, 2017.

\bibitem{syncsynd}
J.~S. Chung and A.~Zisserman, ``Out of time: automated lip sync in the wild,'' in \emph{Asian conference on computer vision}.\hskip 1em plus 0.5em minus 0.4em\relax Springer, 2016, pp. 251--263.

\bibitem{baevski2020wav2vec}
A.~Baevski, Y.~Zhou, A.~Mohamed, and M.~Auli, ``wav2vec 2.0: A framework for self-supervised learning of speech representations,'' \emph{Advances in neural information processing systems}, vol.~33, pp. 12\,449--12\,460, 2020.

\end{thebibliography}
